 \documentclass[pmlr,twocolumn,10pt]{jmlr} 
\usepackage{dblfloatfix}

\usepackage{booktabs}
\usepackage{siunitx}

\usepackage[switch]{lineno}

\theorembodyfont{\upshape}
\theoremheaderfont{\scshape}
\theorempostheader{:}
\theoremsep{\newline}

\jmlrvolume{297}
\jmlryear{2025}
\jmlrworkshop{Machine Learning for Health (ML4H) 2025} 

\usepackage[capitalise,nameinlink,noabbrev]{cleveref}
\crefname{equation}{Eqn.}{Eqns.}
\usepackage{soul}

\usepackage{float}
\usepackage{subcaption}
\usepackage{caption}

\usepackage{bm}
\usepackage{threeparttable}
\usepackage{multirow}
\usepackage{amsmath, amssymb}

\usepackage[table]{xcolor}
\newcommand{\lowcell}[1]{\cellcolor{yellow!10}#1}

\newcommand{\lowwis}[1]
{\cellcolor{blue!5}#1}

 \title[Exploring Time-Step Size in Reinforcement Learning for Sepsis Treatment]{Exploring Time-Step Size in Reinforcement Learning \\ for Sepsis Treatment}

 \author{%
  \Name{Yingchuan Sun} \Email{yingchuan.sun@emory.edu} \\ 
  \Name{Shengpu Tang} \Email{shengpu.tang@emory.edu} \\
  \addr Department of Computer Science, Emory University, USA 
 }

\begin{document}

\maketitle

\begin{abstract}
Existing studies on reinforcement learning (RL) for sepsis management have mostly followed an established problem setup, in which patient data are aggregated into 4-hour time steps. Although concerns have been raised regarding the coarseness of this time-step size, which might distort patient dynamics and lead to suboptimal treatment policies, the extent to which this is a problem in practice remains unexplored. In this work, we conducted empirical experiments for a controlled comparison of four time-step sizes ($\Delta t\!=\!1,2,4,8$ h) on this domain, following an identical offline RL pipeline. To enable a fair comparison across time-step sizes, we designed action re-mapping methods that allow for evaluation of policies on datasets with different time-step sizes, and conducted cross-$\Delta t$ model selections under two policy learning setups. Our goal was to quantify how time-step size influences state representation learning, behavior cloning, policy training, and off-policy evaluation. Our results show that performance trends across $\Delta t$ vary as learning setups change, while policies learned at finer time-step sizes ($\Delta t = 1$ h and $2$ h) using a static behavior policy achieve the overall best performance and stability. Our work highlights time-step size as a core design choice in offline RL for healthcare and provides evidence supporting alternatives beyond the conventional 4-hour setup.

\end{abstract}


\paragraph*{Data and Code Availability}
This study uses the MIMIC-III v1.4 critical care database, which is publicly available to credentialed researchers through PhysioNet. The code for our experiments is available at \url{https://github.com/ysun564/rl4h_timestep}, which builds upon two publicly available code bases.\footnote{\url{https://github.com/microsoft/mimic_sepsis}}\textsuperscript{,}\footnote{\url{https://github.com/MLD3/OfflineRL_FactoredActions}}

\paragraph*{Institutional Review Board (IRB)}
This study does not require IRB approval.
\vspace{-1em}
\section{Introduction}
\label{sec:introduction}
Reinforcement learning (RL) has shown great promise for sequential decision‐making in healthcare, enabling data‐driven treatment policies for complex medical conditions such as sepsis \citep{komorowski2018artificial,tang2024towards,jayaraman2024primer}. Unlike typical RL problems in which states and actions are implicitly assumed to occur at regular intervals, time series data in the electronic health record (EHR) are collected at irregular intervals. This irregularity poses significant challenges for the direct application of RL to such data.

A common workaround is to discretize irregularly sampled data into fixed‐length time steps. For example, in the landmark work by \citet{komorowski2018artificial}, patient data were aggregated into 4-hour time steps. However, it has been demonstrated that this kind of time discretization could introduce biases and obscure rapid physiological changes, negatively impacting policy learning and evaluation \citep{schulam2018discretizingloggedinteractiondata}. So far, this bias has been studied only in theory; nearly all work in this domain has continued to use 4 hours as the time step size and has not systematically studied the impact of other time-step sizes on the entire RL pipeline (see \cref{tab:related_works_abridged}).

In this work, we explore the impact of using four different time-step sizes ($\Delta t\!=\!1,2,4,8$ h) in the MIMIC-III sepsis treatment task. While this may seem to be a simple change in preprocessing, we note that this has important implications for the problem formulation, the study cohort, and the definition of the action space, which pose challenges for establishing a ``fair'' comparison. To facilitate analysis across time-step sizes, we used the same cohort, designed normalized action spaces, and learned and evaluated treatment policies separately for each $\Delta t$, following an identical offline RL pipeline that includes latent state representation learning, behavior cloning, batch-constrained Q-learning (BCQ), hyperparameter selection, and off-policy evaluation (OPE) using weighted importance sampling (WIS) and effective sample size (ESS). To enable a ``fair'' comparison across $\Delta t$ during OPE, we introduce a policy evaluation procedure that uses mapping strategies to transform actions across time-step sizes and evaluate policies learned at a $\Delta t$ on test data that were preprocessed at a different $\Delta t$. We conducted cross-$\Delta t$ model selection for policies trained under two BCQ architectures and evaluated the final selected policies. Our results show that performance trends across $\Delta t$ vary across different BCQ architectures, and finer policies ($t_\pi = 1$ h and $2$ h) trained under BCQ with a static behavior policy tend to exhibit overall good and stable performance. Our work highlights that time-step size is a core design choice for healthcare RL that affects problem formulation, learning and evaluation, and provides empirical evidence for adopting alternatives beyond the conventional 4-hour setup.


\vspace{-0.75em}
\section{Related Work}
\vspace{-0.25em}
When applying RL to ICU sepsis management, most studies discretize each admission's EHR into 4-hour time steps ($\Delta t = 4$ h), and model each interval as a single Markov decision proces (MDP) step. This commonly used design choice is popularized by the ``AI Clinician'' paper \citep{komorowski2018artificial}. In this setting, treatments administered within each 4-hour interval are aggregated into the action, and observations are mapped to a state, forming a trajectory for each admission. In \cref{tab:related_works_abridged} we summarize recent RL for sepsis studies. Nearly all of them adopted $\Delta t=4$ h, inherited from \citet{komorowski2018artificial}. 

\vspace{-0.5em}
\begin{table}[htbp]
\centering
\footnotesize
\caption{Time‐step sizes in prior work that studied RL for sepsis (see \cref{tab:related_works} for full description).}
\label{tab:related_works_abridged}
\scalebox{0.85}{
\begin{tabular}{lc}
\toprule
\textbf{Paper} & $\bm{\Delta t}$ \\
\midrule
\citet{raghu2017continuousstatespacemodelsoptimal} & 4 h \\
\citet{komorowski2018artificial} & 4 h \\
\citet{jeter2019does} & 4 h \\
\citet{8904645} & 1 h \\
\citet{pmlr-v119-tang20c} & 4 h \\
\citet{killian2020empiricalstudyrepresentationlearning} & 4 h \\
\citet{lu2020deepreinforcementlearningready} & 1 h, 4 h \\
\citet{fatemi2021medical} & 4 h \\
\citet{satija2021multi} & 4 h \\
\citet{ji2021trajectory} & 4 h \\
\citet{liang2023treatment} & 4 h \\
\citet{choudhary2024sepsis} & 4 h \\
\citet{tu2025offline} & 1 h \\
\bottomrule
\end{tabular}
}
\end{table}

Whereas most of the studies adopt the 4 h setting, there are also differing viewpoints and attempts regarding the design choice. \citet{jeter2019does} criticizes the coarse discretization for potentially failing to capture rapid physiological changes, thereby providing justification for exploring alternative time-step sizes. \citet{lu2020deepreinforcementlearningready} found that using 1 h time steps significantly altered the learned policy, suggesting that a 4 h step might obscure important decision timing. To our knowledge, no controlled study has been conducted to compare different $\Delta t$ values in otherwise identical setups.

\begin{figure*}[hb]
    \centering
    \includegraphics[width=0.8\linewidth,trim=0 50 0 0]{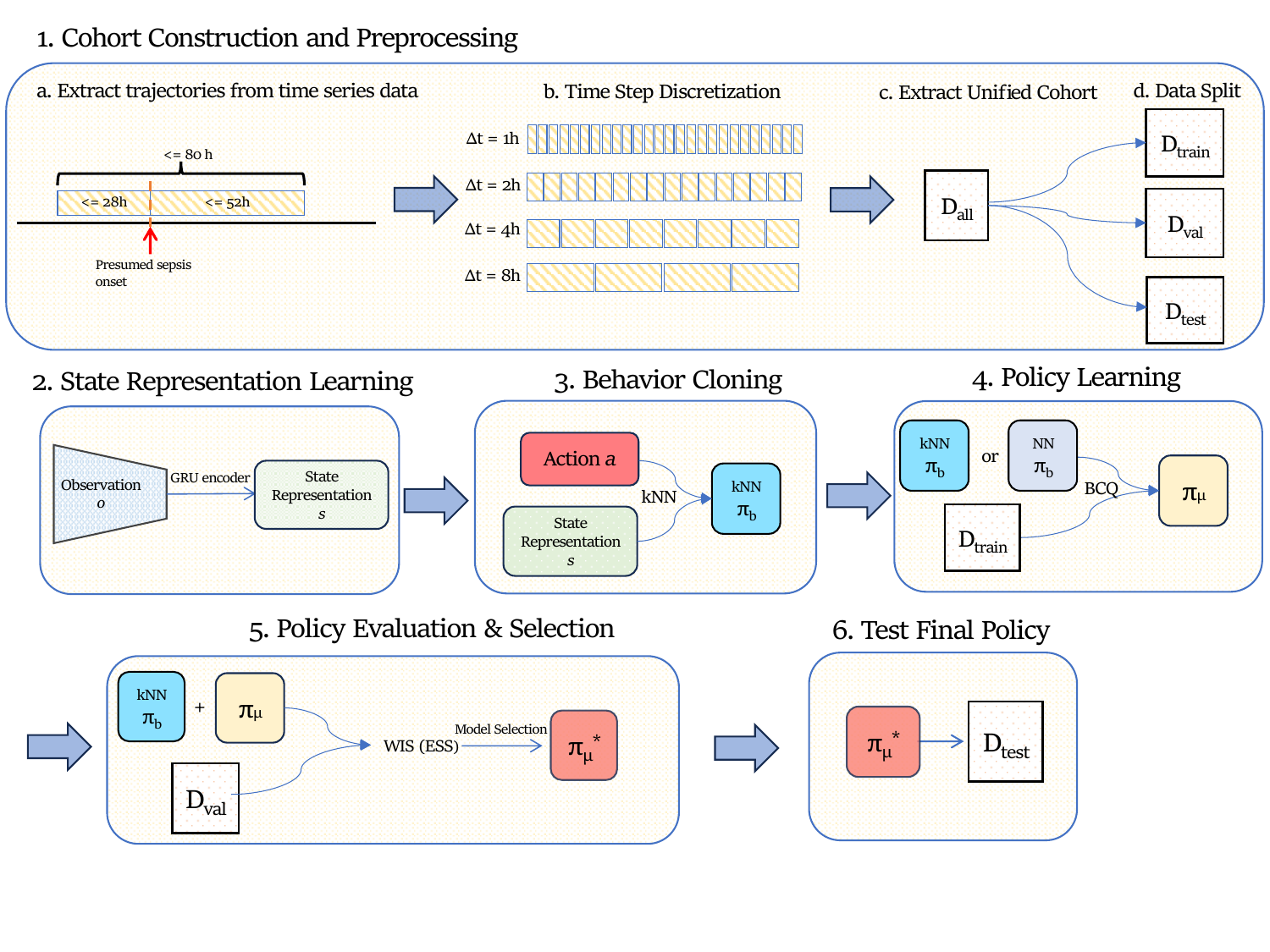}
    \caption{Overview of the offline RL pipeline.}
    \label{fig:offline_rl_overview}
\end{figure*}

\section{Background \& Problem Setup}
\label{sec:formulations}

\subsection{Time Step Discretization}
Suppose the patient timeline starts at an anchor time $t_0$ and ends at an ending time $T$. To convert continuous-time time-series data into a discrete-step trajectory, we discretize the timeline into non-overlapping windows of size $\Delta t$. We define the boundaries between consecutive windows 
\vspace{-0.25em}
\begin{equation*}
t_k = t_0 + k\Delta t,\; k=0,\dots,L, 
\end{equation*}
\vspace{-0.25em}
where \(L = \lceil (T-t_0)/\Delta t\rceil \) represents the total number of time steps.
The \(k\)-th time step is the half-open interval \([t_k,t_{k+1})\) for \(k=0,\dots,T-1\). 


\subsection{Offline RL Objective}

We model sequential clinical treatments as a partially observable Markov decision process (POMDP), defined by the tuple $(\mathcal{S}, \mathcal{A}, \mathcal{O}, P, \Omega, R, \gamma)$, where:
\vspace{-0.5em}
\begin{itemize}\setlength{\itemsep}{0pt}
  \item $\mathcal{S}$ (state space): the set of true patient states, which are latent and unobservable.
  \item $\mathcal{A}$ (action space): the set of possible treatments.
  \item $\mathcal{O}$ (observation space): the set of observable patient measurements (e.g., vitals, labs) per $\Delta t$.
  \item $P$ (transition dynamics): $P(s_{k+1} | s_k,a_k)$ gives the probability of transitioning to next state $s_{k+1}$ from state $s_k$ after action $a_k$.
  \item $\Omega$ (observation function): $\Omega(o_k | s_k)$ gives the probability of observing $o_k$ given the underlying state $s_k$. In our setting, $o_k$ represents all clinical information recorded in the $k$-th time window.
  \item $R$ (reward function): $R(s_k,a_k)$ is the reward obtained after taking action $a_k$ in state $s_k$.
  \item $\gamma$ (discount factor): $\gamma \in [0,1)$ balances immediate and future rewards.
\end{itemize}
\vspace{-0.5em}
In our setting, the true transition dynamics $P(s_{k+1} | s_k,a_k)$ and the observation function $\Omega(o_k | s_k)$ are unknown. We only have access to logged transitions $(o_k, a_k, o_{k+1})$ from offline EHR data. We aggregate the information \textbf{within the window} $[t_k, t_{k+1})$ into $o_k$ and use a learned encoder $f$ to infer a compact latent state from the history of observations, \(s_k = f(o_{0:k})\). 
This latent state $s_k$ serves as the agent’s belief state, and we assume it is a sufficient statistic of history and use it interchangeably with the true state. The treatment action executed \textbf{within the subsequent window} $[t_{k+1},t_{k+2})$ is denoted as $a_k$, selected based on the state $s_k$ following some policy $\pi(a_k|s_k)$, which leads to a reward $r_k$ and influences the transition to the next state $s_{k+1}$ \citep{tang2025off}. The process repeats until a terminal state $s_T$ (e.g., discharge or death) is reached, yielding a trajectory \(\tau = (s_0,a_0,r_0,\dots,s_{T-1},a_{T-1},r_{T-1},s_T)\).
Given a trajectory $\tau$ with rewards $r_0,\ldots,r_{T-1}$, the discounted return is defined as
\(R(\tau) = \sum_{k=0}^{T-1} \gamma^{k} r_k\). 
The goal of RL is to learn an optimal policy $\pi^*$ that maximizes the expected return:
\(\pi^* = {\mathrm{argmax}}_{\pi}\; \mathbb{E}_{\tau \sim \pi} [R(\tau)]\), 
i.e., the policy that achieves the highest expected return. 
In practice, we approximate $\pi^*$ as $\pi_{\mu}$ by applying a learning algorithm to offline data, consisting of trajectories generated by following a behavior policy $\pi_b$. 

\section{Experimental Setup}
\label{sec:experimental_setup}
To empirically study the impact of time step size on the MIMIC sepsis domain, we conducted experiments following an identical offline RL pipeline (\cref{fig:offline_rl_overview}) to data discretized at $\Delta t \in \{1,2,4,8\}$ h, including the following stages: \emph{1.\ Cohort Construction and Preprocessing} $\to$ \emph{2.\ State Representation Learning} $\to$ \emph{3.\ Behavior Cloning} $\to$ \emph{4.\ Policy Learning and Selection} $\to$ \emph{5.\ Policy Evaluation}. Finally, we conducted \emph{6.\ Policy Analysis} to summarize the results of our selected policy.

\subsection{Cohort Construction and Preprocessing}
\label{sec:dataset}
\textbf{Cohort Construction.} 
We used the MIMIC-III v1.4 critical care database \citep{johnson2016mimic}, focusing on adult ICU patients who developed sepsis following the code of \citet{microsoft_mimic_sepsis}. For all patients, we extracted data from their first ICU stay during each hospitalization. Patient data include indicators of infection, patient demographics, and time-series data such as vitals, laboratory results, and interventions (intravenous [IV] fluids, vasopressors, and mechanical ventilation).  Using the Sepsis-3 criteria \citep{singer2016sepsis3}, we identified the presumed onset of infection for each ICU stay.  After that, we assembled each ICU stay’s time series from up to 28 hours before the first sepsis onset to up to 52 hours post-onset, yielding trajectories of up to 80 hours for each ICU admission. 
We discretized each extracted patient trajectory into fixed-length time windows for each time-step size $\Delta t \in \{1,2,4,8\}$. To ensure valid transitions, trajectories shorter than one step at a given $\Delta t$ were excluded. Following \citet{microsoft_mimic_sepsis}, we handled outliers, missing values and implausible measurements, and finally created a separate sepsis cohort for each $\Delta t$. 
Since different numbers of patients were excluded for each $\Delta t$, the resulting cohort sizes were also different. To enable a fair comparison, we defined a unified cohort consisting of ICU stays that were present across all $\Delta t$ cohorts. We then split the cohort into 70/15/15\% for train/validation/test.

\noindent \textbf{POMDP Setup.} For each ICU stay, we extracted $33$ time-varying continuous features per time step, in addition to $5$ static demographic and contextual features (see \cref{tab:features}). Each $38$-dimensional feature vector was considered an observation $o$, while the observation space $\mathcal{O}$ comprises the set of all $o$. Based on this, we learned the (approximate) state space $\mathcal{S}$ as described in \cref{subsec:ais}. Following \citet{komorowski2018artificial}, each action is defined with the interventions applied within each time step, including IV fluids and vasopressors. Specifically, the total volume of IV fluids and the maximum dose of vasopressors applied simultaneously constitute the action taken in that time step. The dosage of fluids and vasopressors was each divided into 5 levels using clinically relevant dosage boundaries \citep{pmlr-v119-tang20c}, yielding an action space $\mathcal{A}$ with $5 \times 5=25$ possible actions. Notably, the dosage levels were normalized by time per $\Delta t$, resulting in a different action space for each $\Delta t$ (\cref{tab:action_space_bins}). We used a sparse reward signal that reflects patient's terminal outcomes \citep{pmlr-v119-tang20c,shi2025between}: a sparse reward of $+100$ was given for survival (at discharge or at end of trajectory) and $0$ otherwise. We set $\gamma = 0.99$ in policy learning and $\gamma = 1$ in policy evaluation \citep{lee2025optimizing}.

\begin{table}[ht]
  \centering
  \caption{Normalized action space $\mathcal{A}$ across $\Delta t$.}
  \label{tab:action_space_bins}
  \centerline{\scalebox{0.8}{
  \begin{tabular}{c cc}
    \toprule
    \textbf{Level} &
    IV fluids (mL/$\Delta t$) & Vasopressors ($\mu$g kg$^{-1}$ min$^{-1}$) \\ \midrule
    0 & $=0$                    & $=0$ \\
    1 & $(0,\, 125\Delta t)$    & $(0,\, 0.08)$ \\
    2 & $[125\Delta t,\, 250\Delta t)$ & $[0.08,\, 0.20)$ \\
    3 & $[250\Delta t,\, 500\Delta t)$ & $[0.20,\, 0.45)$ \\
    4 & $\ge 500\Delta t$       & $\ge 0.45$ \\ \bottomrule
\multicolumn{3}{p{0.66\linewidth}}

\vspace{-3em}
  \end{tabular}
  }}
\end{table}

\subsection{State Representation Learning}
\label{subsec:ais}
To address partial observability in patient trajectories, we learned a compact latent state representation with a recurrent neural network using the approximate information state (AIS) \citep{subramanian2021approximateinformationstateapproximate,killian2020empiricalstudyrepresentationlearning}. The learned state representation constitutes an approximation of the state space $\mathcal{S}$ in the POMDP setup. 
Specifically, we trained a gated recurrent unit (GRU) encoder \citep{cho2014learningphraserepresentationsusing} that, at each step $k$, maps the observations $o_{0:k}$ up to step $k$ and the actions $a_{0:k-1}$ taken up to step $k-1$ to a $D$-dimensional latent state $s_k$. The GRU encoder was optimized via a dual-head objective: one decoder head reconstructs the current observation vector $o_k$, while another head predicts the next observation $o_{k+1}$ given the current latent state $s_k$ and action $a_k$ in the form of a parameterized distribution $p(o_{k+1}|s_k,a_k) = P(s_{k+1} | s_k,a_k) \Omega(o_{k+1}|s_{k+1})$. We trained the representation model on the training set trajectories to minimize the negative log-likelihood (NLL) loss and monitored the learning curve on the validation set.
For each $\Delta t$, we ran an identical grid search over five latent dimension sizes and six different learning rates (see \cref{sec:appendix_A}). The checkpoint with the lowest validation NLL was selected to extract the latent states at each time step. We treat the $D$-dimensional latent state $s_k$ as the AIS summarizing the patient’s history up to time $k$.

\subsection{Behavior Cloning}
\label{sec:bc}
During policy learning and evaluation, we require access to the clinician's action probability distribution. Since the observational dataset only contained the observed deterministic actions, we estimated a stochastic behavior policy $\pi_b$ from data to approximate the clinicians’ non-deterministic treatment decisions. The policy takes the patient’s state representation $s_k$ as input and predicts the clinicians' action distribution $\pi_b(a|s_k)$. 
After considering discriminative performance and calibration, we implemented $k$-nearest neighbors (kNN) classifiers for behavior cloning separately for each data partition (train/validation/test) \citep{raghu2018behaviour}. We transformed the episodic dataset into a flattened dataset of $n$ state-action pairs $(s_k, a_k)$, where each sample corresponds to one step in a trajectory. Each state vector $s_k$ is treated as a feature input, and its associated action $a_k$ is the label used for kNN classification. We then performed a hyperparameter grid search over the number of neighbors $k$ and the distance metric (see \cref{tab:hyper_search}). Best classifiers were selected based on their macro and micro averaged area under the receiver operating characteristic curve (AUROC) via 5-fold cross validation, and were used as the $\pi_b$ for BCQ and OPE. We report the result of grid search and policy performance in \cref{sec:bc_results}. 

\vspace{-0.8em}
\subsection{Policy Learning}

In healthcare where exploration of new treatments is infeasible, it is critical that we do not learn a policy that extrapolates to actions (more specifically, state-action combinations) not observed in the data \citep{article}. To address the issue, we used an offline RL algorithm, namely (discrete-action) batch-constrained Q-learning (BCQ) \citep{fujimoto2019offpolicydeepreinforcementlearning}. 
In our BCQ implementation, the Q-network is a three-layer feed-forward network that estimates $Q(s, a)$, together with a target network of identical architecture updated via Polyak averaging. At each update, the Q-network selects the action for the next state from a set generated by $\pi_b$ of the behavior cloning stage, where actions whose estimated behavior probability falls below a threshold $\varepsilon$ are masked out. The target network then evaluates the selected action when forming the bootstrapping target. We trained the Q-network with the Huber loss between the current and target values. 
In classic BCQ implementations, behavior policy $\pi_b$ is modeled by a neural network and is typically trained concurrently with the Q-network \citep{liu2022avoiding, tang2023leveragingfactoredactionspaces}. To ensure a consistent $\pi_b$ throughout the policy training and evaluation stages, we used the static kNN-based $\pi_b$ in BCQ (the resulting policies are called \textbf{kNN-policies}). For comparison, we also considered a standard neural network for $\pi_b$ in BCQ training (the resulting policies are called \textbf{NN-policies}). 
For each $\Delta t$, we ran experiments with both BCQ architectures by training for a fixed number of epochs and conducted a grid search using five different random seeds and eight values of $\varepsilon$ (see \cref{sec:appendix_A}), which yielded a set of learned policies $\pi_\mu$. 

\vspace{-0.8em}
\subsection{Policy Evaluation \& Selection}
\noindent\textbf{Off-policy Evaluation (OPE).} 
We evaluated the performance of the learned policy using OPE, specifically weighted importance sampling (WIS). 
The standard WIS estimator used importance weights to reweight the returns of test trajectories under the assumption that test data were generated by the behavior policy $\pi_b$; using our learned $\pi_b$, we computed cumulative per-step importance ratios $w_i$ 
for each action $a_k$ taken by clinicians, and then took a weighted average of the observed returns $G_i$ normalized by the sum of the importance weights across all evaluation trajectories (\cref{eq:wis}) \citep{10.5555/2969033.2969163}. 

\vspace{-1.5em}
\begin{subequations}\label{eq:wis}
\begin{align}
w_i = \prod_{k=0}^{T_i-1} \frac{\pi_{\mu}(a_{i,k}\mid s_{i,k})}{\pi_b(a_{i,k}\mid s_{i,k})}, &\quad G_i = \sum_{k=0}^{T_i-1} \gamma^{k} r_{i,k} \label{eq:wis-defs} 
\end{align}

\begin{equation}
\widehat{V}_{\mathrm{WIS}} = \frac{\sum_{i=1}^{n} w_i\, G_i}{\sum_{i=1}^{n} w_i} \label{eq:wis-main} \\
\end{equation}
\end{subequations}

\noindent To handle trajectories with different lengths, we implemented the per-horizon WIS estimator \citep{10.5555/3304652.3304742}, which normalizes the cumulative importance weights separately at each time step $k$ over trajectories that survive to $k$. We note that the same patient trajectory has a larger $H$ when discretized at a smaller $\Delta t$ and this directly increases estimator variance. In order to make results comparable across $\Delta t$, we truncated the cumulative importance ratios \(W \;=\; \prod_{k=1}^{H} \rho_k \) at $W \le 1.438^{H}$ \citep{3b3cebfc-9984-3188-9eff-75a75713a089}.
As WIS requires the action distribution $\pi_{\mu}(a_k| s_k)$ of treatment policy $\pi_{\mu}$ to calculate importance ratio $\rho$, we applied $\epsilon$-softening to convert deterministic greedy action recommended by the policy into stochastic policy probabilities \citep{tang2021model}: $\tilde{\pi}(a| s) = (1-\varepsilon)\,\mathbf{1}\{a=a^\star\}
\;+\; \varepsilon\,\hat{\pi}_b(a| s)$ where the greedy action is $a^\star=\arg\max_a Q(s,a)$. 
This avoids zero weights in WIS and thereby prevents ESS from becoming too small. Throughout the experiments, we use $\epsilon = 0.1$.  

To quantify the reliability and variance of WIS, we also recorded the effective sample size (ESS) of WIS \citep{elvira2022rethinking}:
\begin{equation}\label{eq:ess}
\mathrm{ESS} = \frac{\bigl(\sum_{i=1}^{n} w_i\bigr)^2}{\sum_{i=1}^{n} w_i^2},
\end{equation}
which reflects how many trajectories contribute meaningfully after weighting. 
Although we have two BCQ architectures that use different behavior policies $\pi_b$ for training, for OPE we only present results for kNN-based $\pi_b$ learned in \cref{sec:bc} as NN-based $\pi_b$ yielded low ESS.

\vspace{0.25em}
\noindent\textbf{Cross-$\Delta t$ OPE.} 
By default, the policy learned at a particular time-step size $\Delta t$ will be evaluated on test datasets processed at the same $\Delta t$. This poses a challenge for directly comparing policies learned at different $\Delta t$ since they would be effectively evaluated on a ``different test set''. 
To allow for a fairer comparison, we evaluated policy learned at some $\Delta t = t_\pi$ using a test dataset with a different $\Delta t = t_D$. This includes two cases, where (i) $t_\pi > t_D$ and (ii) $t_\pi < t_D$. 
We achieve this by defining how to map the $t_\pi$ action taken by the policy to the $t_D$ action observed in data. For simplicity, we assume that $t_\pi$ and $t_D$ are integer multiples of each other, such that either (i) $t_{\pi} = M t_D$ or (ii) $N t_{\pi} = t_D$ for integers $M,N$. 
\vspace{-0.5em}
\begingroup
\addtolength{\leftmargini}{-0.6em}

\begin{itemize}\setlength{\itemsep}{0pt}
\item \textbf{Mapping when $t_\pi > t_D$.} Here, each action taken by the policy spans a time interval of size $t_{\pi}$, which corresponds to $M = t_\pi/t_D$ intervals of size $t_{D}$. Conceptually, for each action taken by the policy at $t_\pi$, we broadcast that action over the entire interval, yielding a sequence of $M$ identical $t_D$ actions (see case 1 in \cref{fig:mapping}). Because the action space is normalized across $\Delta t$, broadcasting does not change the action index.
\vspace{-0.5em}

\item \textbf{Mapping when $t_\pi < t_D$.} In this case, each $t_D$ interval corresponds to a sequence of $N = t_D/t_\pi$ finer steps. 
However, as the data has been discretized at $t_D$ hours, the intermediate states that would be observed at the $N$ fine steps (which are $t_\pi$ hours apart) are unobserved. Thus, within each $t_D$ window, the policy can only predict an action for the first $t_\pi$ interval. In this case, we repeat that action over all $N$ fine steps and aggregate it to a single $t_D$ action using mapping rules below (see case 2 in \cref{fig:mapping}). 
For IV fluids, we used an \textbf{expected-overlap rule}. We first compute the total fluid volume interval $[L, U]$ by summing the upper and lower bounds (\cref{tab:action_space_bins}) of the $N$ fluid actions at $t_{\pi}$, and then identify the fluid volume interval at $t_D$ with the largest overlap. Since fluid volume level-4 does not specify an upper bound, we used the 95\textsuperscript{th}-percentile empirical threshold on the training set (for each $\Delta t$). For vasopressors, we take the maximum level across the $N$ fine steps. The indices of fluids and vasopressors combined yield a joint $t_D$ action (\cref{tab:action_space_bins}). 

\end{itemize}
\endgroup
\vspace{-0.3em}


\noindent\textbf{Model Selection.}
For each $(t_\pi, t_D)$ pair where $t_\pi, t_D \in \{1, 2, 4, 8\}$ h under two BCQ architectures, we conduct OPE via per-horizon WIS and mapping rules. In the model selection stage, we present the validation ESS–WIS Pareto frontier for candidate policies, which consists of the set of candidate policies for which no other policy simultaneously achieves both higher WIS and higher ESS.
We picked the policy on each frontier that balanced both metrics, together with an ESS cutoff that constrains minimum value for ESS. This approach prevents the selection of models that achieve high WIS but with excessive variance. We select one policy for each $(t_\pi, t_D)$ pair, resulting in $4^2 = 16$ policies for each BCQ architecture.

\subsection{Policy Analysis}
For the selected policies, we report WIS and ESS on the test set with bootstrapped confidence intervals (CIs), and present results separately for each $(t_\pi, t_D)$ pair. We compared performance of policies with different $t_\pi$ using dataset with the same $t_D$ to ensure a fair comparison (i.e., using an identical test set). To eliminate variations inherent to the WIS estimator itself, we additionally report OPE results measured using fitted-Q evaluation (FQE). To complement these metrics, we further include heatmaps in \cref{sec:appendix_B} showing how the learned policies redistribute action probabilities relative to the clinician policy.

\vspace{-1em}
\section{Results}
We applied our experimental pipeline to four time-step sizes ($ \Delta t\!=\!1,2,4,8$ h) under two BCQ architectures, then conducted OPE for each $(t_\pi, t_D)$ pair. For each $\Delta t$\, we report the following: (i) cohort statistics, (ii) AIS reconstruction error across latent dimensions, (iii) BC performance, and (iv) performances of the kNN-policies and NN-policies learned by two BCQ architectures, measured by per-horizon WIS and action frequency heatmaps. 

\vspace{-1em}
\subsection{Cohort Statistics}
\label{subsec:cohort_stats}
In \cref{subsec:cohort_size}, we compare the cohort sizes across $\Delta t$. The cohort sizes decrease with coarser $\Delta t$, reflecting the exclusion of trajectories shorter than one step. For all experiments, we report results on a unified cohort (\cref{tab:cohort}) that includes trajectories present at all $\Delta t$, which contains 18,377 ICU stays with a mortality rate of 5.9\%. 

\begin{table}[ht]
  \centering
  \caption{Cohort statistics across $\Delta t$.}
  \label{tab:cohort}

  \resizebox{0.48\textwidth}{!}{%
    \begin{tabular}{lccc}
      \toprule
      & N & \% Female  & Mean ICU Hours \\
      \midrule
      Survivors      & 17{,}288 & 44.2\% & 45.7 \\
      Non-survivors  &  1{,}089 & 44.6\% & 61.4 \\
      \bottomrule
    \end{tabular}
  }
\end{table}

\vspace{-1em}
\subsection{State Representation Learning}
Following \cref{subsec:ais}, we first evaluated the quality of our AIS encoder across time-step sizes. \cref{tab:ais-mse-latent} reports the selected latent dimension, learning rate and the resulting minimum validation mean squared error (MSE) with 95\% confidence intervals (CI). 
The same latent dimension (128) was selected for all $\Delta t$. 
For the learning rate, 0.0001 was selected for $\Delta t = 8 h$, whereas 0.001 was selected for the other $\Delta t$. We also observe that validation MSE tends to increase with larger $\Delta t$. This is likely because the task is inherently more difficult for longer prediction horizons, as the AIS encoder can be seen as a forecaster that predicts future observations with a prediction horizon of $\Delta t$ (e.g., average heart rate over the next time step).
\begin{table}[t]
  \centering
  \caption{AIS encoder (GRU) results across time-step sizes: selected latent dimension, learning rate, and minimum validation MSE with 95\% confidence intervals from 1000 bootstrap samples.}
  \label{tab:ais-mse-latent}
  \vspace{-0.75em}
  \resizebox{\linewidth}{!}{
    \begin{tabular}{cccc}
      \toprule
      $\Delta t$ (h) & latent dim & learning rate & MSE [95\% CI] \\
      \midrule
      1 & 128 & 0.001 & 0.2288 [0.2181, 0.2424] \\
      2 & 128 & 0.001 & 0.2678 [0.2655, 0.2702] \\
      4 & 128 & 0.001 & 0.4011 [0.3940, 0.4110] \\
      8 & 128 & 0.0001 & 0.4356 [0.4290, 0.4426] \\
      \bottomrule
    \end{tabular}
  }
\end{table}

\vspace{-0.75em}
\subsection{Behavior Cloning}
\begin{table}[t]
  \centering
  \caption{Estimated performance (Macro and Micro AUROC) of kNN behavior policy on the validation sets across time-step sizes, with 95\% confidence intervals from 1000 bootstrap samples.}
  \label{tab:bc}
  \vspace{-0.75em}
  \small
  \resizebox{\linewidth}{!}{
  \begin{tabular}{ccc}
    \toprule
    $\Delta t$ (h) & Macro AUROC [95\% CI] & Micro AUROC [95\% CI] \\
    \midrule
    1 & 0.7715 [0.7678, 0.7753] & 0.9449 [0.9443, 0.9456] \\
    2 & 0.8047 [0.7998, 0.8095] & 0.9491 [0.9482, 0.9500] \\
    4 & 0.8143 [0.8071, 0.8211] & 0.9507 [0.9496, 0.9518] \\
    8 & 0.7754 [0.7601, 0.7883] & 0.9483 [0.9466, 0.9501] \\
    \bottomrule
  \end{tabular}
  }
\end{table}

\label{sec:bc_results}

Across all settings, as $k$ increases from $21$ to $5\sqrt{n}$, the cross-validation macro- and micro-AUROC generally improve. Based on the cross-validation performance, we selected kNN classifiers for each data partition with Euclidean distance and $k=5\sqrt{n}$ as $\pi_b$, yielding macro-AUROC $>0.75$ and micro-AUROC $\approx 0.95$. We summarize the performance in \cref{tab:bc}. While class imbalance can reduce macro-AUROC and inflate micro-AUROC, the overall performance is comparable to past work \citep{jeong2024identifyingdifferentialpatientcare}.

\subsection{Policy Learning, Evaluation \& Selection}
\label{subsec:ope}

\begin{table*}[ht]
\centering
\small
\caption{Cross-$\Delta t$ evaluation results with 95\% CI for selected kNN-Policies (left) and NN-Policies (right).}
\label{tab:cross-dt}
\setlength{\tabcolsep}{4pt}
\centerline{\scalebox{0.8}{
\begin{tabular}{
cc
S[table-format=2.2] l S[table-format=3.2] l S[table-format=2.2] l
S[table-format=2.2] l S[table-format=3.2] l S[table-format=2.2] l
}
\toprule
& & \multicolumn{6}{c}{\textbf{kNN-Policy}} & \multicolumn{6}{c}{\textbf{NN-Policy}}\\
\cmidrule(lr){3-8}\cmidrule(lr){9-14}
$t_D$ & $t_\pi$
& \multicolumn{2}{c}{PHWIS} & \multicolumn{2}{c}{ESS} & \multicolumn{2}{c}{FQE}
& \multicolumn{2}{c}{PHWIS} & \multicolumn{2}{c}{ESS} & \multicolumn{2}{c}{FQE}\\
\cmidrule(lr){3-4}\cmidrule(lr){5-6}\cmidrule(lr){7-8}
\cmidrule(lr){9-10}\cmidrule(lr){11-12}\cmidrule(lr){13-14}
 &  & {Test} & \multicolumn{1}{c}{CI} & {Test} & \multicolumn{1}{c}{CI} & {Test} & \multicolumn{1}{c}{CI}
 & {Test} & \multicolumn{1}{c}{CI} & {Test} & \multicolumn{1}{c}{CI} & {Test} & \multicolumn{1}{c}{CI} \\
\midrule
\multirow{4}{*}{1}
 & \textbf{1} & 95.82 & {[93.47, 97.90]} & 47.74 & {[38.00, 59.66]} & 94.38 & {[94.69, 94.90]}
 & 95.69 & {[94.19, 96.83]} & 160.31 & {[136.36, 181.49]} & 96.38 & {[96.38, 96.55]} \\
 & 2          & 96.26 & {[94.12, 97.98]} & 46.61 & {[35.25, 60.63]} & 96.87 & {[96.87, 97.05]}
 & \lowwis{93.83$^{\downarrow}$} & {[90.76, 96.96]} & 68.25 & {[52.10, 86.72]} & \lowwis{92.22$^{\downarrow}$} & {[92.07, 92.29]} \\
 & 4          & 97.58 & {[94.03, 99.39]} & \lowcell{25.65$^{\dagger}$} & {[18.03, 36.72]} & 94.11 & {[93.77, 94.03]}
 & \lowwis{90.62$^{\downarrow}$} & {[87.40, 95.23]} & \lowcell{30.14$^{\dagger}$} & {[21.96, 40.36]} & 96.43 & {[96.27, 96.46]} \\
 & 8          & \lowwis{93.33$^{\downarrow}$} & {[91.60, 96.42]} & 49.61 & {[37.34, 61.57]} & \lowwis{93.38$^{\downarrow}$} & {[92.57, 92.79]}
 & 96.35 & {[93.75, 98.92]} & \lowcell{15.99$^{\dagger}$} & {[9.29, 23.77]} & 96.19 & {[95.74, 95.91]} \\
\addlinespace
\multirow{4}{*}{2}
 & 1          & \lowwis{93.48$^{\downarrow}$} & {[91.07, 95.77]} & 73.76 & {[60.81, 87.40]} & 95.30 & {[95.11, 95.43]}
 & \lowwis{93.45$^{\downarrow}$} & {[90.82, 98.77]} & \lowcell{21.07$^{\dagger}$} & {[13.28, 29.43]} & 94.91 & {[94.69, 94.98]} \\
 & \textbf{2} & 95.16 & {[92.64, 97.11]} & 81.45 & {[65.75, 95.39]} & 95.43 & {[94.98, 95.30]}
 & \lowwis{93.71$^{\downarrow}$} & {[91.28, 96.28]} & 84.51 & {[71.47, 99.00]} & 95.77 & {[95.17, 95.46]} \\
 & 4          & \lowwis{91.35$^{\downarrow}$} & {[85.93, 95.94]} & \lowcell{45.07$^{\dagger}$} & {[34.37, 56.14]} & 95.43 & {[95.03, 95.30]}
 & \lowwis{91.14$^{\downarrow}$} & {[87.09, 96.06]} & \lowcell{21.83$^{\dagger}$} & {[14.16, 32.26]} & 95.47 & {[95.58, 95.84]} \\
 & 8          & \lowwis{91.80$^{\downarrow}$} & {[88.74, 95.36]} & 80.92 & {[64.06, 93.96]} & 95.16 & {[95.36, 95.60]}
 & 96.88 & {[91.58, 99.89]} & \lowcell{2.19$^{\dagger}$} & {[1.07, 5.28]} & 94.55 & {[94.37, 94.64]} \\
\addlinespace
\multirow{4}{*}{4}
 & 1          & 94.87 & {[92.66, 96.90]} & 163.28 & {[146.77, 184.35]} & 95.30 & {[95.40, 95.66]}
 & 96.44 & {[92.14, 99.48]} & \lowcell{16.05$^{\dagger}$} & {[12.47, 24.27]} & \lowwis{93.13$^{\downarrow}$} & {[93.04, 93.35]} \\
 & 2          & 94.87 & {[92.65, 96.80]} & 163.28 & {[143.44, 183.71]} & 95.30 & {[95.37, 95.61]}
 & 95.43 & {[92.45, 98.33]} & 62.35 & {[52.69, 76.50]} & \lowwis{92.91$^{\downarrow}$} & {[92.86, 93.10]} \\
 & \textbf{4} & 94.92 & {[92.53, 96.73]} & 161.81 & {[145.37, 181.07]} & 95.30 & {[95.37, 95.66]}
 & \lowwis{94.01$^{\downarrow}$} & {[91.23, 97.15]} & 66.61 & {[56.10, 77.72]} & 94.15 & {[93.69, 94.10]} \\
 & 8          & 94.37 & {[92.02, 96.30]} & 144.43 & {[127.35, 163.27]} & 95.30 & {[95.40, 95.69]}
 & 96.48 & {[90.71, 99.72]} & \lowcell{18.72$^{\dagger}$} & {[13.46, 25.35]} & 94.66 & {[94.56, 94.87]} \\
\addlinespace
\multirow{4}{*}{8}
 & 1          & 94.12 & {[92.16, 95.70]} & 340.91 & {[320.00, 367.78]} & 96.66 & {[96.49, 96.97]}
 & \lowwis{93.55$^{\downarrow}$} & {[88.41, 98.54]} & \lowcell{48.78$^{\dagger}$} & {[40.66, 60.13]} & \lowwis{92.59$^{\downarrow}$} & {[92.40, 92.69]} \\
 & 2          & 94.13 & {[92.48, 96.00]} & 326.67 & {[304.63, 353.73]} & 96.66 & {[96.47, 96.87]}
 & \lowwis{91.71$^{\downarrow}$} & {[87.74, 95.80]} & \lowcell{135.41$^{\dagger}$} & {[123.96, 151.25]} & 96.06 & {[95.93, 96.35]} \\
 & 4          & \lowwis{92.66$^{\downarrow}$} & {[90.72, 94.78]} & 311.25 & {[287.34, 330.94]} & 96.66 & {[96.48, 96.93]}
 & 97.08 & {[95.39, 98.58]} & 213.58 & {[190.51, 232.46]} & 97.78 & {[97.62, 98.00]} \\
 & \textbf{8} & \lowwis{94.06$^{\downarrow}$} & {[92.00, 95.71]} & 305.91 & {[288.45, 326.83]} & 96.66 & {[96.47, 96.88]}
 & 94.12 & {[92.43, 95.60]} & 431.44 & {[405.32, 454.87]} & 95.32 & {[94.93, 95.35]} \\
\bottomrule
\end{tabular}
}}
\vspace{3pt}
\footnotesize $^{\downarrow}$ Performance lower than the clinician baseline (94.09). $\dagger$ Low ESS relative to the corresponding ESS cutoff.
\end{table*}






\begin{figure}
\vspace{-2em}
\centering
\begin{tabular}{c}
    \includegraphics[width=0.75\linewidth]{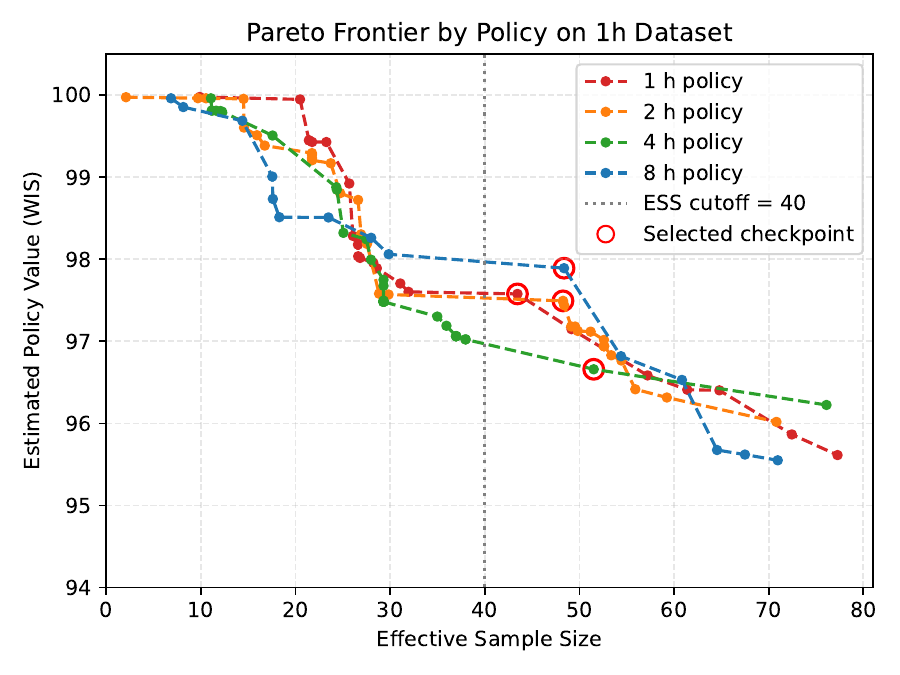} \\
    \includegraphics[width=0.75\linewidth]{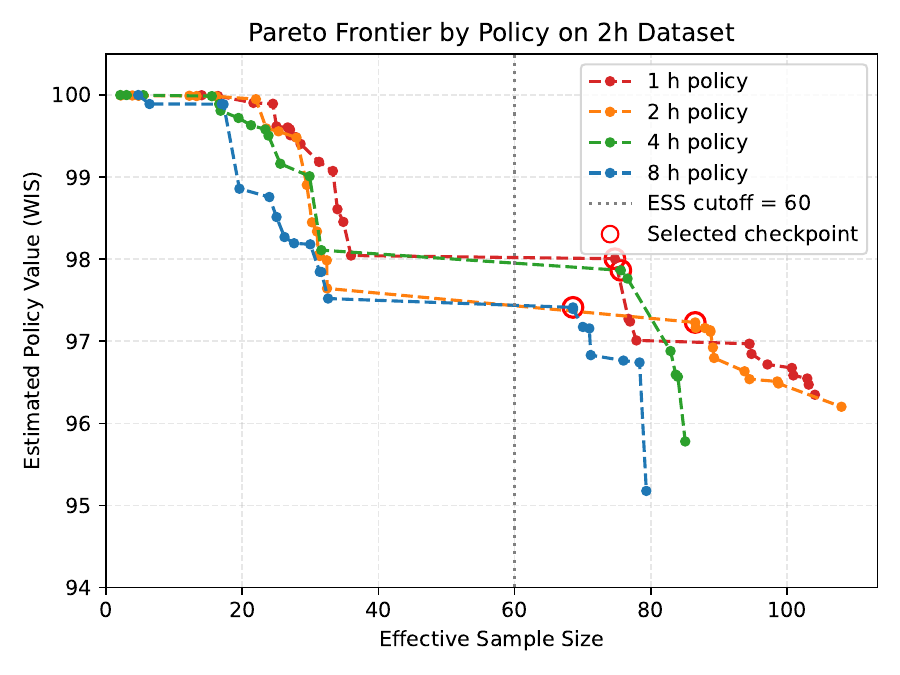} \\
    \includegraphics[width=0.75\linewidth]{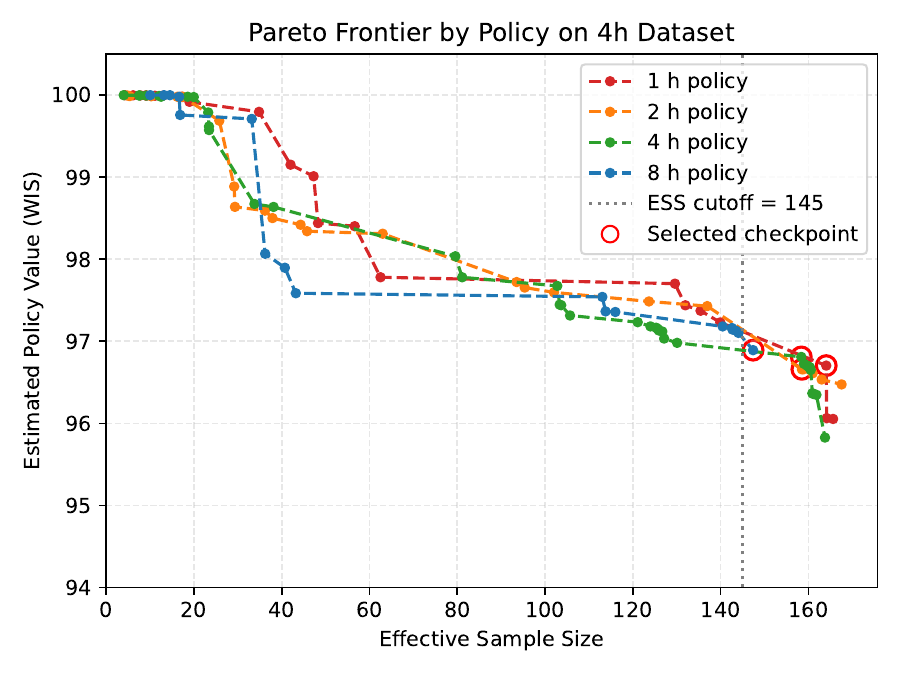} \\
    \includegraphics[width=0.75\linewidth]{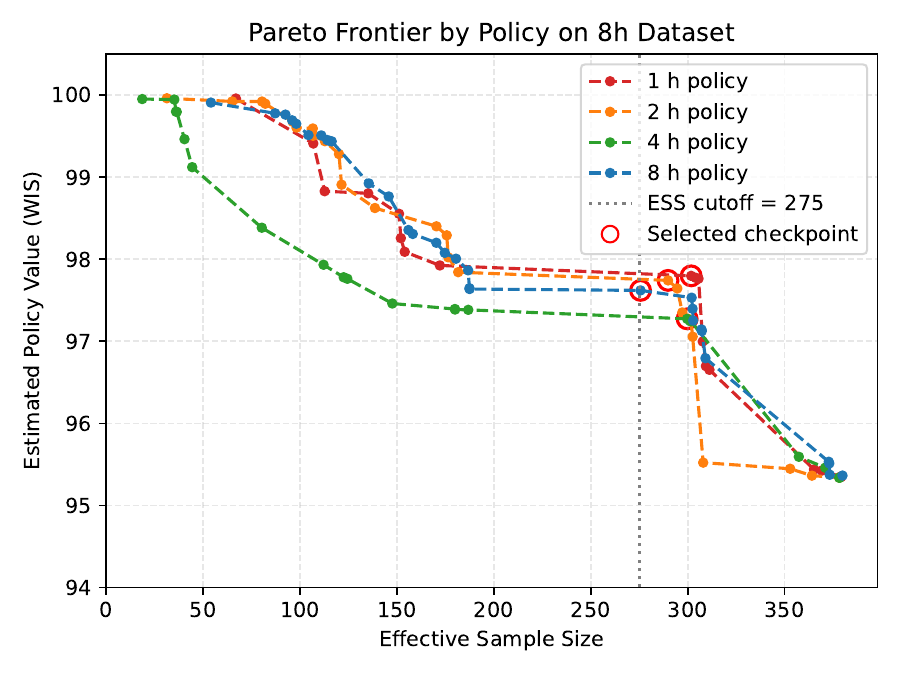} \\
\end{tabular}
\caption{kNN-policies' Pareto frontiers of performance (WIS vs ESS) across evaluation $t_d$. Each curve corresponds to policies trained at $t_\pi$; hollow markers denote the model selected for testing; dotted lines with different colors represent the thresholds used as the boundary for model selection across $\Delta t$.}
\label{fig:pareto_kNN}
\vspace{-1em}
\end{figure}

\cref{fig:pareto_kNN,fig:pareto_NN} show validation ESS–WIS Pareto frontiers for candidate kNN/NN policies across $t_D \in \{1, 2, 4, 8\}$ h.
For all figures, a similar trend appears: WIS is close to 100 when ESS is small, and then declines as ESS increases, reflecting the trade-off between high value and high confidence. In general, evaluation using validation data at a larger $t_D$ led to a wider range of ESS (up to $>$350 for 8 h) compared to a smaller $t_D$ (up to 80 for 1 h). This is because a finer $t_D$ yields a POMDP with more decision points, giving the evaluation policy more opportunities to diverge from the behavior policy and thereby increasing the variance of the importance weights. 

All Pareto figures of the kNN-policies (\cref{fig:pareto_kNN}) show a visually similar shape, where the curves for $t_\pi \in \{1, 2, 4, 8\}$ h exhibit substantial overlap, and no single $t_\pi$ policy curve consistently dominates the others. All curves achieve validation WIS results higher than that of the clinicians (94.09). For model selection, we selected a different ESS cutoff of 40/60/145/275 respectively for each $t_D \in \{1, 2, 4, 8\}$ h, and chose the policy that achieved the highest WIS and has an ESS $\geq$ the cutoff. 

The Pareto figures of the NN-policies (\cref{fig:pareto_NN}) noticeably differ in shape from those of the kNN-policies. When the policies are learned at the same time-step size as validation data ($t_\pi = t_D$), the corresponding curves generally have the widest ESS range compared to policies learned at a different $\Delta t$. We set a ESS cutoff of 75/70/60/300 respectively for each $t_D \in \{1, 2, 4, 8\}$ h to ensure comparable variance of the selected policies learned at different $t_{\pi}$.

\subsection{Test Performance}

\label{subsec:test}
\cref{tab:cross-dt} summarizes final policy performance on the test set, while \cref{fig:heatmap_kNN,fig:heatmap_NN} show action frequency heatmaps of the final kNN and NN policies. We summarize the key findings below.

\noindent \textbf{\textit{How do performance trends differ for kNN- vs.\ NN-policies?}} 
Overall, we find that the kNN-policies are more stable than NN-policies. 
The ESS values of kNN-policies are similar within each $t_D$ and almost always exceed the corresponding $t_D$-specific ESS cutoff with rare exceptions (2 out of 16 total cases). 
In contrast, the ESS for NN-policies drops below the cutoff in 9 cases. 
Interestingly, the highest ESS of NN-policies for each $t_D$ always occurs when $t_\pi = t_D$. For example, when $t_\pi = t_D = 1$h, the NN-policy achieves the highest ESS of 160.31. 
In terms of policy value as measured by PHWIS and FQE, kNN-policies exhibit fewer cases than NN-policies (7 vs.\ 12 out of 32 cases) where the policy value is lower than the clinician baseline, indicating overall better performance. In summary, kNN-policies tend to have generally better performance and stability than NN-policies. 
Therefore, we focus our subsequent results on kNN-policies.

\noindent \textbf{\textit{Which $t_{\pi}$ has the best performance?}} 
When $t_D = 1$ h or $2$ h, the finer policies ($t_\pi = 1, 2$ h) tend to perform better; they achieve better WIS and FQE scores than the $t_\pi = 8$ h policy with a comparable ESS. This may be because policies learned at a coarser timescale struggle to accurately predict actions at a more granular timescale, leading to higher variance (as seen for $t_\pi = 4$ h) and degraded performance (as for $t_\pi = 8$ h). 
Among policies learned at a finer timescale, the $t_\pi = 2$ h policy achieves higher values in both WIS and FQE than the $t_\pi = 1$ h policy, possibly reflecting a balance between time granularity and performance stability.

When $t_D = 2$ h and $4$ h, the finer policies ($t_\pi = 1, 2$ h) tend to perform similarly with coarser ones.
With $t_\pi = 1$ h or $2$ h, the kNN-policies at $t_D = 4$ h show WIS and ESS values close to those of the $t_\pi = 4$ h policy, with identical FQE scores, indicating similar policy behaviors after the cross-$\Delta t$ mapping. This is also reflected in the corresponding heatmaps (see \cref{fig:heatmap_kNN}).
At $t_D = 8$ h, the $t_\pi = 1$ h and $2$ h kNN-policies show slightly higher WIS and ESS values than the $t_\pi = 4$ h policy, while their FQE scores remain the same. 

Overall, we found that kNN-policies learned at finer time-step sizes ($\Delta t = 1, 2$ h) were the most stable and performant across the evaluation settings. 

\vspace{-1em}
\section{Conclusion \& Discussion}
\vspace{-0.5em}

While nearly all prior work on RL for sepsis has universally followed the AI Clinician \citep{komorowski2018artificial} with 4 h time steps, this work presents, to our knowledge, the first systematic comparison across four different time-step sizes (1, 2, 4, 8 h) under two policy training setup (BCQ with kNN and NN behavior cloning) using an identical offline RL pipeline that includes preprocessing, representation learning, behavior cloning, policy learning, and off-policy evaluation. 
To enable a fair comparison of different time step sizes beyond simply altering preprocessing, we controlled for several aspects of our pipeline. We used the same cohort and a fixed set split across all settings, preventing the impact of data differences on training and evaluation. We designed a normalized action space for each $\Delta t$ to facilitate comparison. We conducted AIS grid searches and selected the best latent dimension independently for each $\Delta t$. We tuned BC policies, selecting different $k$ for each $\Delta t$ to ensure downstream stability. For OPE, we clipped importance ratios based on each trajectory's horizon, thereby mitigating effects arising solely from differing trajectory lengths resulting from differing time-step sizes. Together, these choices provide a robust reference for future studies that intend to conduct similar investigations on this domain.

Furthermore, we introduced a method for cross-$\Delta t$ policy evaluation. We developed action mappings for IV fluids and vasopressors based on their definitions and conducted model selection on all $t_\pi$ policies across all $t_D$ datasets. Our results show that finer policies ($t_\pi = 1, 2$ h) trained under BCQ with a kNN behavior policy tend to exhibit overall good and stable performance. These findings underscore that time-step size fundamentally changes the task, thus shaping learned policies and the evaluation process. 

Still, our work has several limitations and challenges. During cohort construction stage, in order to build a unified cohort for direct comparison across $\Delta t$, we only included admissions that are present in \textit{all} four initial cohorts (\cref{tab:cohort_size} vs. \cref{tab:cohort}). This reduced the amount of data available in the experiment and might introduce selection bias since we only kept trajectories that span at least 8 hours, which may in turn affect the performance of the learned policies. In the behavior cloning stage, there is no widely accepted standard for measuring the quality of the estimated behavior policy. Although we followed prior studies and used AUROC as the metric for hyperparameter selection during behavior cloning, future work should explore how other behavior cloning method and hyperparameter selection metrics may influence policy learning and evaluation. In the evaluation stage, we introduced a cross-$\Delta t$ evaluation method. However, this approach is based on our current action space design and may not directly be applicable to a more general setting. In future work, we plan to investigate alternative mapping strategies that apply more broadly to different types of action spaces. In addition, the choice of the ESS cutoffs in our model selection stage was manually determined, as we currently lack a standardized criterion for specific $\Delta t$. This process might introduce human bias and skew the results. Future work should explore and design fairer model selection methods both theoretically and empirically. There also remain challenges in interpreting the final results. Although many of our learned policies outperform the clinician baseline on the test set in terms of the evaluation metrics, the heatmap action distributions differ substantially from those of clinicians. This discrepancy may limit the clinical utility of the learned policies. While wes believe that having a robust understanding of technical differences across $\Delta t$ is a crucial step before potential real-life use, future work should aim to strengthen the clinical validation of the approach.

Our results demonstrate that time-step size is a crucial design choice for clinical RL tasks that can substantially shape the learned policies. Our work advocates for careful reconsideration from the community of different time-step sizes in sepsis management beyond the conventional 4 h setup, in order to learn better policies and enable fairer evaluation across time-step sizes.

\section*{Acknowledgments}
This work was supported in part by computing resources provided by the Department of Computer Science at Emory University. 
The authors thank members of the TAIL group (Aayam Kc, Yuxuan Shi, Matthew Lafrance, Sixing Wu, Canyu Cheng, Hanqi Chen, Yilang Ding, Kaixuan Liu, Tina Piltner, Owen Tucker, Olin Gilster, Zixin Lin) for helpful discussions regarding this work. We also appreciate the constructive feedback from the anonymous reviewers of ML4H 2025, NeurIPS 2025 TS4H workshop and ARLET workshop, and RLC 2025 RL4RS workshop. 
%

\bibliography{refs}
\clearpage
\onecolumn

\appendix

\section{Extended Methods}
\label{sec:appendix_A}

\label{subsec:cohort_size}
\begin{table}[H]
  \centering
  \caption{Extracted cohort size of MIMIC-Sepsis at different time steps.}
  \label{tab:cohort_size}
  \begin{tabular}{lc}
    \toprule
    $\Delta t$ (h) & Cohort Size \\
    \midrule
    1 & 18,995 \\
    2 & 18,987 \\
    4 & 18,906 \\
    8 & 18,783 \\
    \bottomrule
  \end{tabular}
\end{table}

\phantomsection\label{app:papers}
\begin{table*}[htbp]

{\centering
\caption{RL studies for sepsis care, summarizing time‐step choices and key design aspects.}
\label{tab:related_works}
\centerline{\scalebox{0.75}{
\begin{tabular}{@{ }p{4cm} c p{3.6cm} p{2cm} c p{7cm}@{}}
\toprule
\textbf{Paper} & $\bm{\Delta t}$ & \textbf{Algorithm} & \textbf{Dataset} & \textbf{Cohort} & \textbf{Notes} \\
\midrule
\citet{raghu2017continuousstatespacemodelsoptimal} & 4 h & Dueling DDQN & MIMIC-III & 17.9k & Continuous state; 5×5 IV/vaso bins; first DL-RL policy (–3.6 \% mortality). \\[0.5ex]

\citet{komorowski2018artificial} & 4 h & Batch Q-learning & MIMIC-III (+eRI$^{*}$) & 17.1k & AI Clinician; 750 states, 25 actions; external validation. \\[0.5ex]

\citet{jeter2019does} & 4 h & Reproduction study & MIMIC-III & 5.4k & Finds no-action policy often rivals AI Clinician; urges caution. \\[0.5ex]
    
\citet{8904645} & 1 h & Deep IRL & MIMIC-III & 14.0k & Learns reward; highlights mortality factors (e.g.\ PaO$_2$). \\[0.5ex]

\citet{pmlr-v119-tang20c} & 4 h & Set-valued DQN & MIMIC-III & 20.9k & Returns top-$k$ near-optimal dose sets for clinician choice. \\[0.5ex]

\citet{killian2020empiricalstudyrepresentationlearning} & 4 h & Offline DQN & MIMIC-III & 17.9k & Sequential latent encodings outperform raw features. \\[0.5ex]

\citet{lu2020deepreinforcementlearningready} & 1 h, 4 h & Dueling DDQN & MIMIC-III & 17k+ & Sensitivity study on features, reward, time discretization. \\[0.5ex]

\citet{fatemi2021medical} & 4 h & Dead-end discovery & MIMIC-III & 17k+ & Identifies high-risk states; secures policy to avoid them. \\[0.5ex]

\citet{satija2021multi} & 4 h & MO-SPIBB & MIMIC-III & 17k+ & Safe policy improvement under performance constraints. \\[0.5ex]

\citet{ji2021trajectory} & 4 h & Trajectory inspection & MIMIC-III & 17k+ & Clinician “what-if” review reveals policy flaws; validation tool. \\[0.5ex]

\citet{liang2023treatment} & 4 h & Episodic-memory DQN & MIMIC-III & 17.9k & Memory module boosts sample efficiency, lowers est.\ mortality. \\[0.5ex]

\citet{choudhary2024sepsis} & 4 h & Tabular MDP & MIMIC-III & $\sim$18k & ICU-Sepsis benchmark: 715 states, 25 actions. \\[0.5ex]

\citet{tu2025offline} & 1 h & CQL (offline) & MIMIC-III & 14.0k & Safety-aware CQL with dense rewards for variable-length stays. \\[0.5ex]

\bottomrule
\end{tabular}
}}
}
\vspace{0.5em}

{\footnotesize
$^{*}$eRI: Philips eICU Research Institute cohort for external validation;
DDQN: Double Deep Q-Network;  
DQN: Deep Q-Network;  
IRL: Inverse Reinforcement Learning;  
CQL: Conservative Q-Learning;  
MO-SPIBB: Multi-Objective Safe Policy Improvement with Baseline Bootstrapping.
}
\end{table*}

\label{app:features}
\begin{table*}[h]
  \centering
  \caption{Observed features extracted from the MIMIC-III database. 
           The upper panel lists the 33-dimensional time-varying continuous variables fed to the GRU encoder, 
           following the default code configuration. 
           The lower panel lists the 5 static demographic / contextual variables appended to each trajectory.}
  \label{tab:features}

  \begin{tabular}{|p{0.29\linewidth}|p{0.29\linewidth}|p{0.29\linewidth}|}
    \multicolumn{3}{c}{\textbf{33-d Time-varying continuous features}}\\[0.25em]\hline
    Glasgow Coma Scale     & Heart Rate        & Sys.\ BP\\
    Dia.\ BP               & Mean BP           & Respiratory Rate\\
    Body Temp (℃)          & FiO\textsubscript{2} & Potassium\\
    Sodium                 & Chloride          & Glucose\\
    INR                    & Magnesium         & Calcium\\
    Hemoglobin             & White Blood Cells & Platelets\\
    PTT                    & PT                & Arterial pH\\
    Lactate                & PaO\textsubscript{2} & PaCO\textsubscript{2}\\
    PaO\textsubscript{2}/FiO\textsubscript{2} & Bicarbonate (HCO\textsubscript{3}) & SpO\textsubscript{2}\\
    BUN                    & Creatinine        & SGOT\\
    SGPT                   & Bilirubin         & Base Excess\\\hline
  \end{tabular}

  \vspace{1em}
  \begin{tabular}{|p{0.95\linewidth}|}
    \multicolumn{1}{c}{\textbf{5-d Demographic and contextual features}}\\[0.2em] \hline
    Age \quad\textbullet\quad
    Gender \quad\textbullet\quad
    Weight\quad\textbullet\quad
    Ventilation Status \quad\textbullet\quad
    Re-admission Status  \\ \hline
  \end{tabular}
\end{table*}

\begin{table}[htbp]
  \centering
  \caption{Hyperparameter values used for training GRU encoder and BCQ models.}
  \label{tab:hyper_search}
  \vspace{0.4em}
  \begin{threeparttable}
  \begin{tabular}{@{}l c@{}}
    \toprule
    \textbf{Hyperparameter} & \textbf{Searched Settings} \\
    \midrule
    \multicolumn{2}{@{}l}{\textbf{RNN:}}\\
    \quad– Embedding dimension, $d_S$ & $\{8,16,32,64,128\}$ \\
    \quad– Learning rate              & $\{1\!\times\!10^{-5},\,3\!\times\!10^{-5},\,1\!\times\!10^{-4},\,3\!\times\!10^{-4},\,5\!\times\!10^{-4},\,1\!\times\!10^{-3}\}$ \\
    \midrule
    \multicolumn{2}{@{}l}{\textbf{kNN:}}\\
    \quad– Number of neighbors, $k$ 
      & $k_i = \exp \ \!\bigl(\ln 21 + \tfrac{i}{7}(\ln(5\sqrt{n}) - \ln 21)\bigr)$\tnote{a} \\
      
    \quad– Distance metric            
      & $\{\text{Euclidean, Manhattan}\}$ \\
    \midrule
    \multicolumn{2}{@{}l}{\textbf{BCQ (with 5 random restarts):}}\\
    \quad– Threshold, $\varepsilon$  & $\{0,0.01,0.05,0.1,0.3,0.5,0.75,0.999\}$ \\
    \quad– Learning rate              & $3\!\times\!10^{-4}$ \\
    \quad– Weight decay               & $1\!\times\!10^{-3}$ \\
    \quad– Hidden layer size          & $256$ \\
    \bottomrule
  \end{tabular}
  \begin{tablenotes}
    \item[a] $i=0,1,\dots,7$. $n$ denotes the size of the flattened dataset.
  \end{tablenotes}
  \end{threeparttable}
\end{table}

\begin{figure*}[t]
    \centering
\includegraphics[width=1.2\linewidth,trim=0 0 0 50]{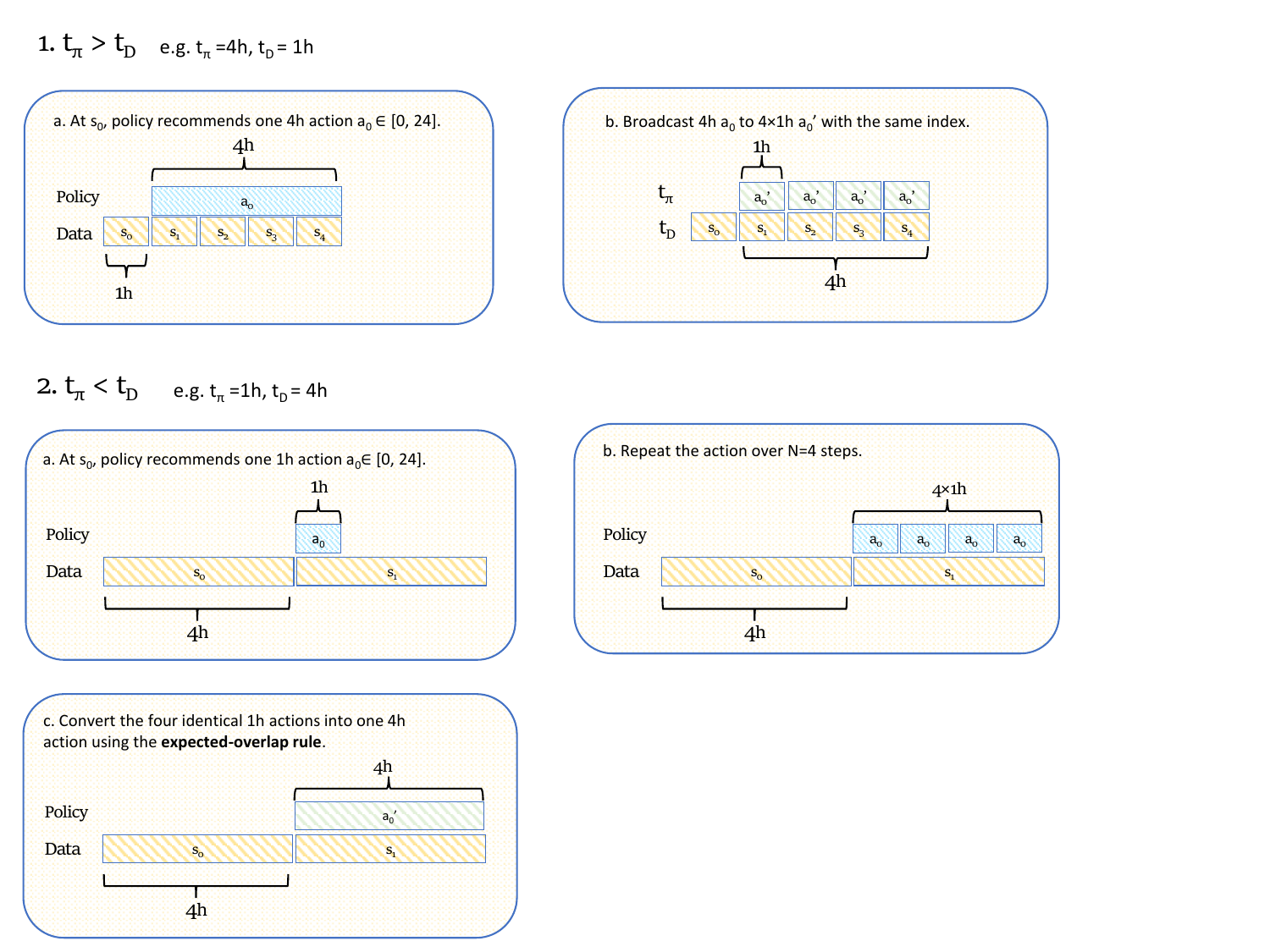}
    \caption{The illustration of our cross-$\Delta t$ mapping.}
    \label{fig:mapping}
\end{figure*}

\vspace{5em}
\clearpage
\section{Extended Results}
\label{sec:appendix_B}
\begin{figure}[htbp]
\centering
\vspace{-1em}
\begin{minipage}[t]{.48\linewidth}
  \includegraphics[width=\linewidth]{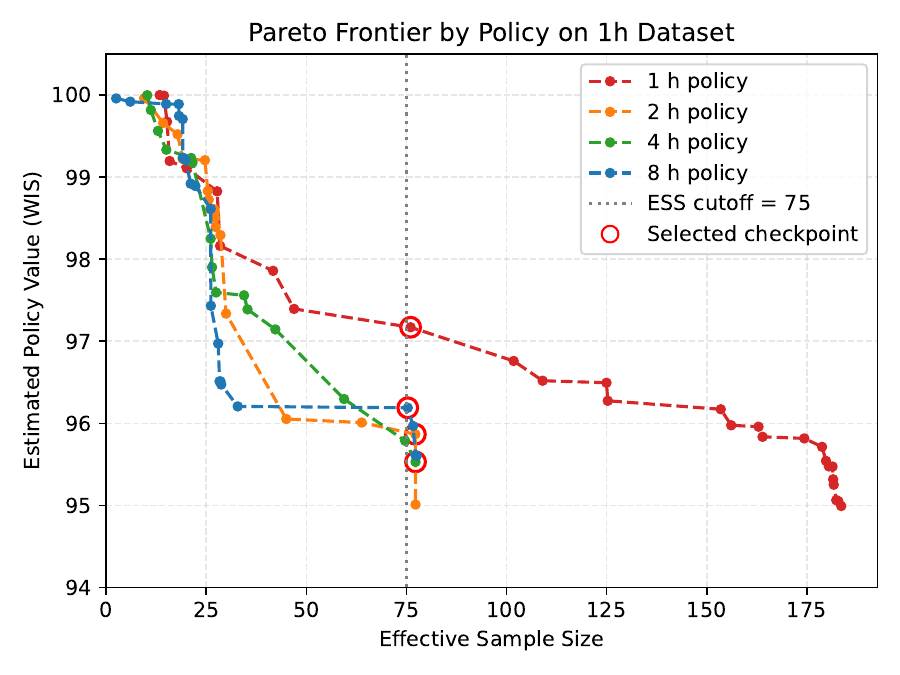}
  \caption*{$t_D=1\,\mathrm{h}$ dataset}
\end{minipage}
\hfill
\begin{minipage}[t]{.48\linewidth}
  \includegraphics[width=\linewidth]{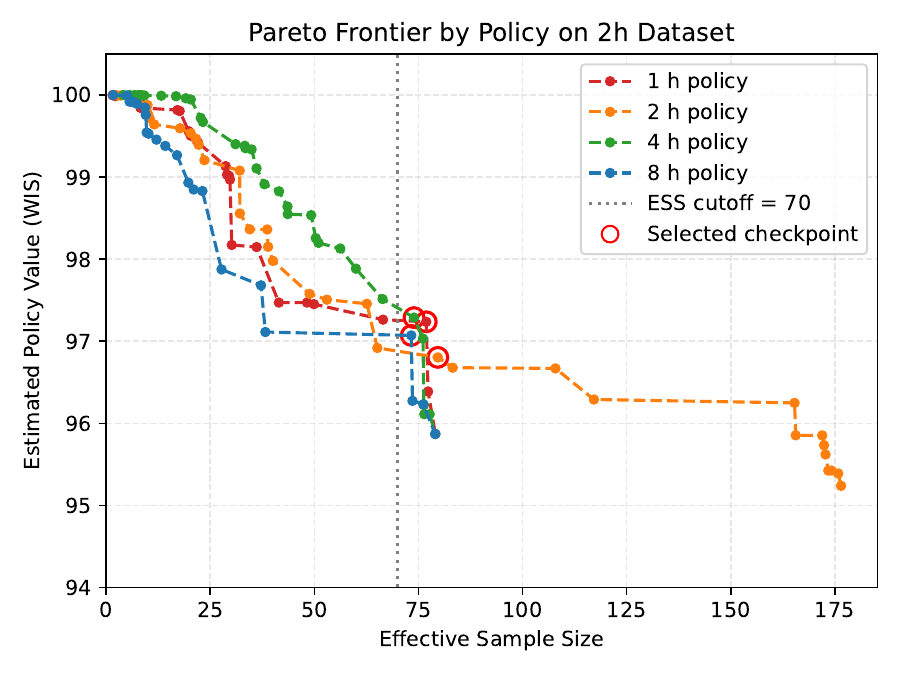}
  \caption*{$t_D=2\,\mathrm{h}$ dataset}
\end{minipage}

\par\medskip

\begin{minipage}[t]{.48\linewidth}
  \includegraphics[width=\linewidth]{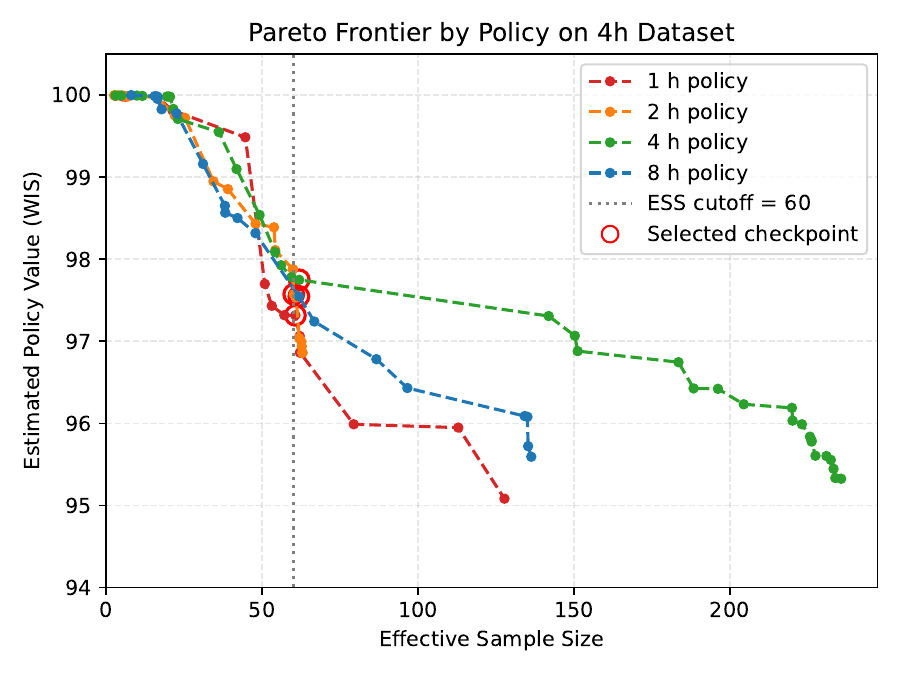}
  \caption*{$t_D=4\,\mathrm{h}$ dataset}
\end{minipage}
\hfill
\begin{minipage}[t]{.48\linewidth}
  \includegraphics[width=\linewidth]{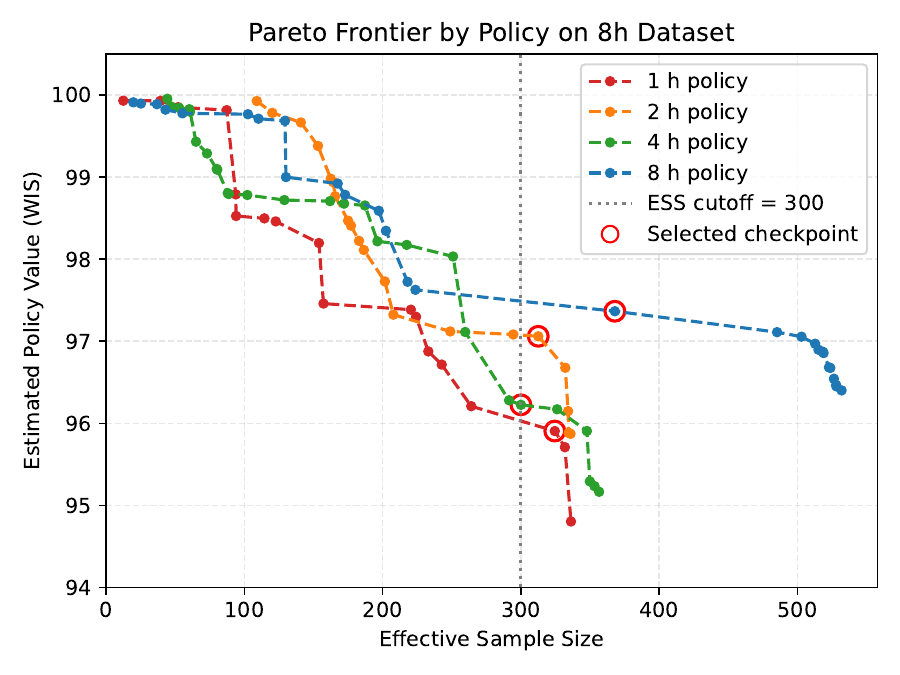}
  \caption*{$t_D=8\,\mathrm{h}$ dataset}
\end{minipage}

\caption{NN-policies' pareto frontiers of performance (WIS vs.\ ESS) across evaluation time steps $t_D$. Each curve corresponds to a policy trained at a specific $t_\pi$; hollow markers denote the model selected for testing; dotted lines with different colors represents the thresholds used as the boundary for model selection across $\Delta t$.}
\label{fig:pareto_NN}
\end{figure}

\begin{figure*}[t]
  \centering

  \includegraphics[width=\textwidth]{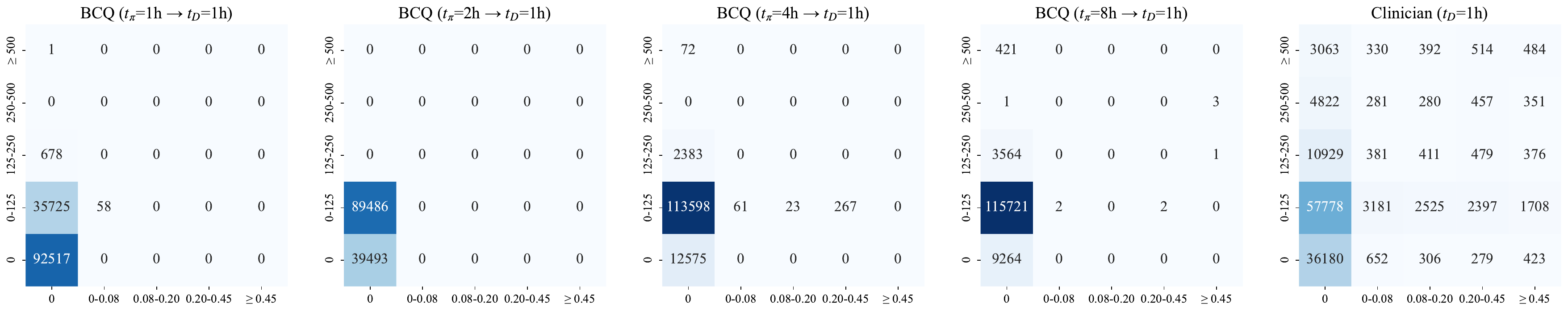}
\caption*{$t_D=1\,\mathrm{h}$} 
\vspace{1.5em}

  \includegraphics[width=\textwidth]{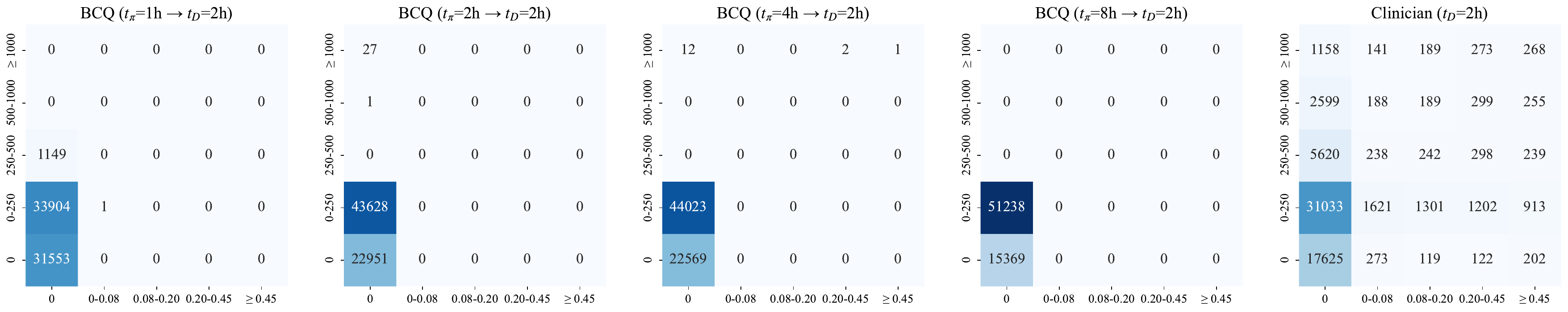}
\caption*{$t_D=2\,\mathrm{h}$}
\vspace{1.5em}

  \includegraphics[width=\textwidth]{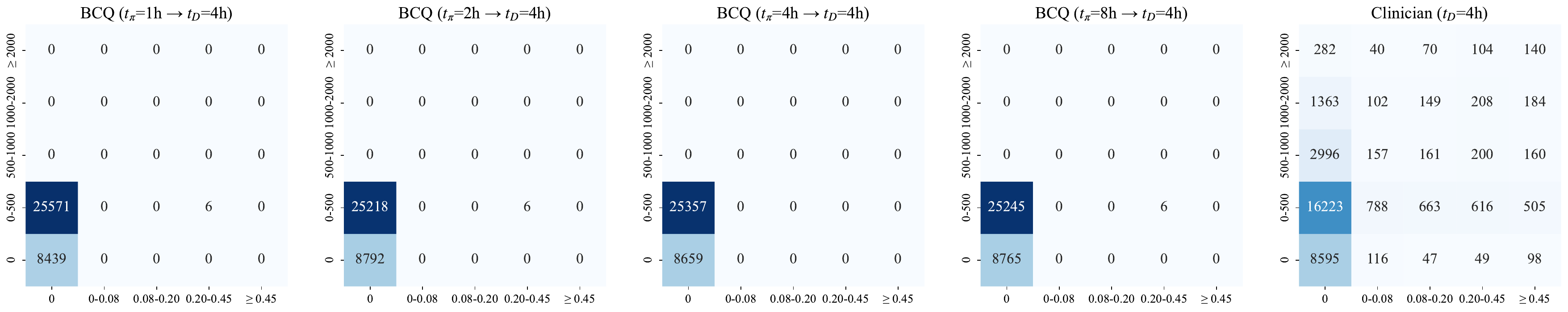}
\caption*{$t_D=4\,\mathrm{h}$}
\vspace{1.5em}

  \includegraphics[width=\textwidth]{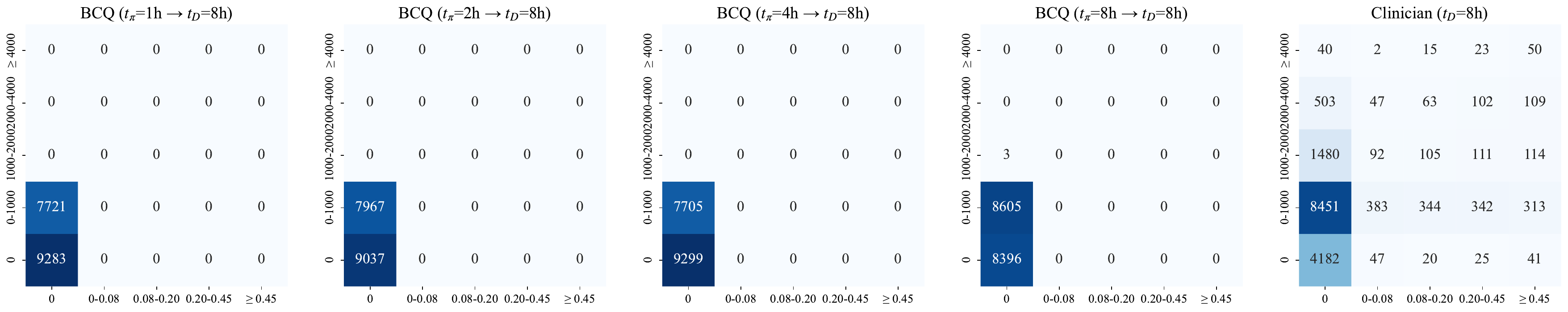}
\caption*{$t_D=8\,\mathrm{h}$}
  \caption{kNN policies' frequency heatmap of IV fluids (y-axis; mL) and vasopressors (x-axis;
$\mu\mathrm{g}\,\mathrm{kg}^{-1}\,\mathrm{min}^{-1}$) doses on validation set. The columns from left to right represent respectively: Policies at $t_\pi \in \{1, 2, 4, 8\}$ h, clinician policy. Darker cells indicate more frequent selections. Almost all policies most frequently select actions with zero vasopressor and low IV fluids doses.}
  \label{fig:heatmap_kNN}
\end{figure*}

\begin{figure*}[t]
  \centering

  \includegraphics[width=\textwidth]{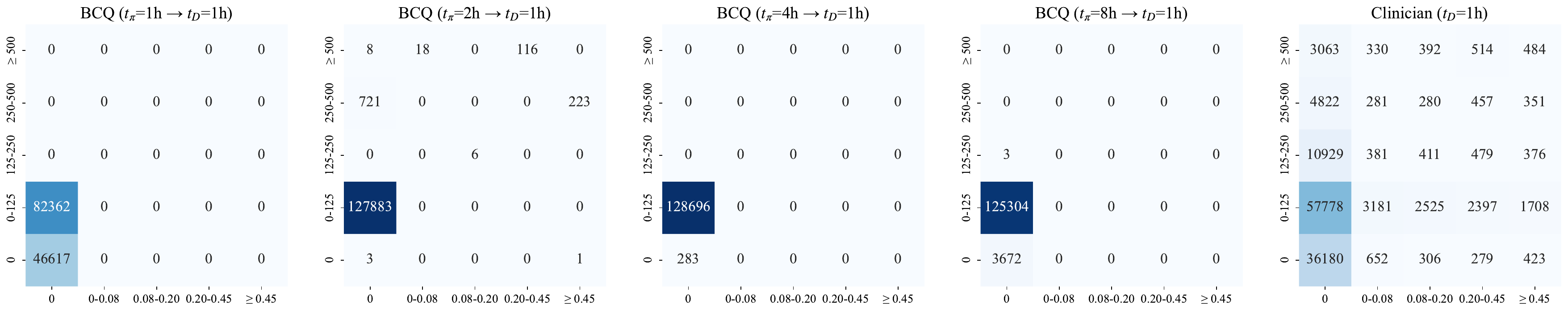}
\caption*{$t_D=1\,\mathrm{h}$}
\vspace{1.5em}

  \includegraphics[width=\textwidth]{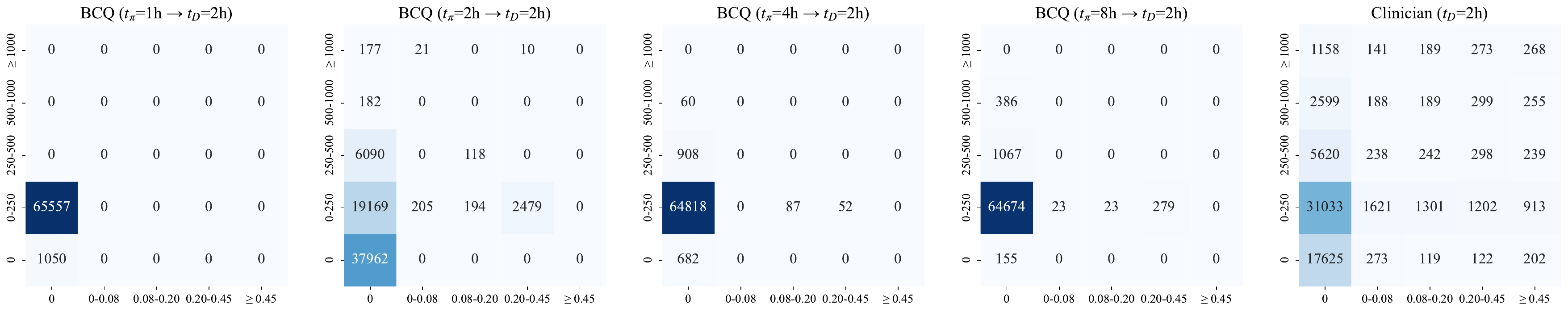}
\caption*{$t_D=2\,\mathrm{h}$}
\vspace{1.5em}

  \includegraphics[width=\textwidth]{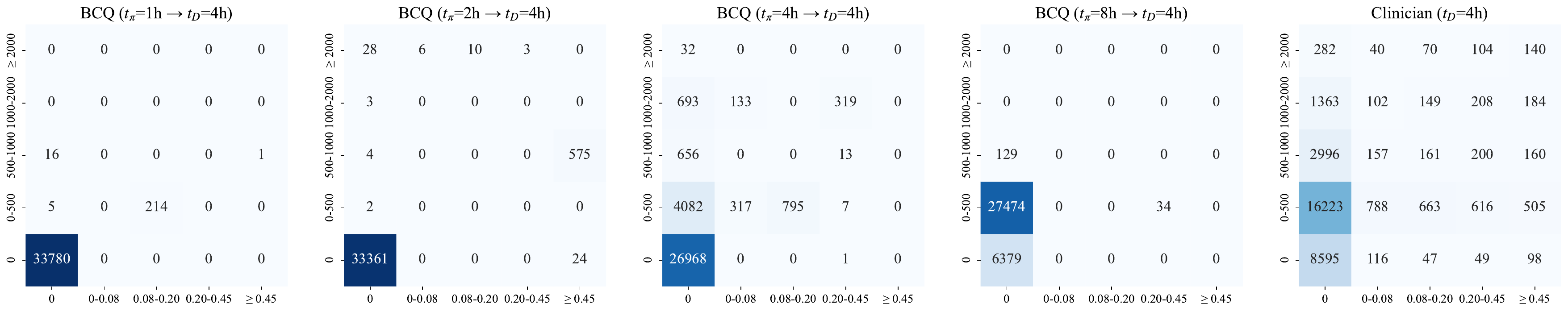}
\caption*{$t_D=4\,\mathrm{h}$}
\vspace{1.5em}

  \includegraphics[width=\textwidth]{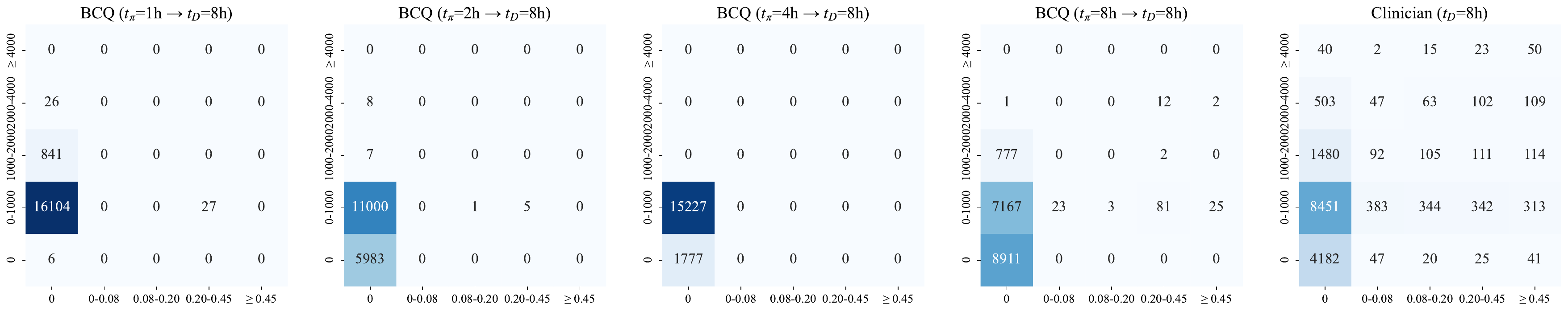}
\caption*{$t_D=8\,\mathrm{h}$}
  \caption{NN policies' frequency heatmap of IV fluids (y-axis; mL) and vasopressors (x-axis;
$\mu\mathrm{g}\,\mathrm{kg}^{-1}\,\mathrm{min}^{-1}$) doses on validation set. The columns from left to right represent respectively: policies at $t_\pi \in  \{1, 2, 4, 8\}$ h, clinician policy. Darker cells indicate more frequent selections. Almost all policies most frequently select actions with zero vasopressor and low IV fluids doses.}
  \label{fig:heatmap_NN}
\end{figure*}

    
        

        


\end{document}